\documentclass[11pt,a4paper]{article}
\usepackage[hyperref]{acl2020}
\usepackage{times}
\usepackage{latexsym}
\usepackage{graphicx}
\usepackage{amsmath}
\usepackage{xspace}

\usepackage{microtype}

\aclfinalcopy % Uncomment this line for the final submission
\newcommand{\Sref}[1]{\S\ref{#1}}

\newcommand{\Fref}[1]{Figure~\ref{#1}}

\newcommand{\Tref}[1]{Table~\ref{#1}}

\newcommand\BertTrans{\textsc{Trans}\xspace}
\newcommand\BertRnn{\textsc{Rnn}\xspace}
\newcommand\BertRnnSc{\textsc{Rnn+Sc}\xspace}
\newcommand\BertRnnAttn{\textsc{Rnn+Attn}\xspace}

\title{A Generative Approach to Titling and Clustering Wikipedia Sections}

\author{Anjalie Field \\
  Carnegie Mellon University\thanks{\;\;Work done while the first author was an intern at Google Research.} \\
  \texttt{anjalief@cs.cmu.edu} \\\And
  Sascha Rothe \\
  Google Research \\
  \texttt{rothe@google.com} \\\And
  Simon Baumgartner \\
  Google Research \\
  \texttt{simonba@google.com} \\\AND
   Cong Yu \\
   Google Research \\
   \texttt{congyu@google.com} \\\And
   Abe Ittycheriah \\
   Google Research \\
   \texttt{aittycheriah@google.com} }

\date{}

\begin{document}
\maketitle

\begin{abstract}
We evaluate the performance of transformer encoders with various decoders for information organization through a new task: generation of section headings for Wikipedia articles. Our analysis shows that decoders containing attention mechanisms over the encoder output achieve high-scoring results by generating extractive text. In contrast, a decoder without attention better facilitates semantic encoding and can be used to generate section embeddings. We additionally introduce a new loss function, which further encourages the decoder to generate high-quality embeddings.
\end{abstract}

\section{Introduction}

Automated information labeling and organization has become a desirable way to process the copious amounts of available text. We develop methods for producing text headings and section-level embeddings through a new task: generation of section titles for Wikipedia articles. This task is useful for improving Wikipedia, an active area of research due to the long tail of poor quality articles, including articles lacking section subdivisions or consistent headings \citep{lebret2016neural,piccardi2018structuring,liu2018generating}. Additionally, the types of labels used to denote sections can be useful for organizing other unstructured collections of text. 

We approach this task in two ways: first we train a text generation model for producing section titles, and second, we leverage our model architecture to extract section embeddings, which offer a useful mechanism for comparing and clustering sections with similar information \citep{banerjee2007clustering,hu2009exploiting,reimers2019classification}. This approach provides a flexible framework for creating paragraph-level embeddings, in which the type of information encoded in the embedding can be controlled by changing the generation task.

Section title generation is similar to existing tasks, such as generating titles for newspaper articles \cite{rush2015neural,nallapati2016abstractive}. However, Wikipedia section titles contain a unique mix of short abstractive headings like ``History'' and longer extractive headings like song titles, where many of the words in the section title also appear in the section text. The variations in the type of headings makes this dataset useful for analyzing how models perform on different subsets of the data. 

A common state-of-the-art model for many existing text generation tasks uses an encoder-decoder framework where the encoder is initialized with BERT and the decoder is also a transformer \cite{vaswani2017attention,devlin2019bert,zhang2019pretraining,rothe2019leveraging}. The entire output of the encoder is passed to the decoder, which allows the decoder to attend over the entire input sequence during each generation step.

In contrast, we explore using transformer encoders with RNN decoders and show that RNN decoders better generate short abstractive titles while transformer decoders perform better on longer extractive titles. Embeddings extracted from the RNN decoders also perform better in clustering evaluations, which suggests that the attention-based mechanisms in the transformer facilitate copying input text into the output, but the RNN architecture better facilitates encoding semantic meaning. 

We additionally introduce a new loss function for the RNN decoder that encourages the start and end states of the RNN to be similar. This loss function encourages the model to encode meaningful information into a single state, which further improves the quality of the generated section-level embeddings. 

We first describe our models (Section~\ref{sec:model}) and our data set  (Section~\ref{sec:data}) and then present results, evaluating our models on a held-out test corpus (Section~\ref{sec:results}). Our main contributions include: (1) the introduction of a new short-text generation task that is useful for information labeling and organization; (2) an analysis of text generation models for this task; (3) the introduction of a novel loss function that results in  high-quality section embeddings.

\section{Models}
\label{sec:model}
Our primary task is to generate section titles, and our secondary task is to generate section-level embeddings. All models use an encoder-decoder architecture, where the encoder is initialized with BERT \cite{devlin2019bert}. We use 4 decoder variants, including one trained with a novel loss function.

\textbf{\BertTrans} This model contains a (randomly initialized) transformer decoder, with hyperparameters identical to the BERT-base model. The hidden states generated by the encoder for the entire input sequence are passed to the decoder, thus allowing the decoder to attend over the entire input sequence during each decoding step. This model serves as our primary baseline, as it is identical to the BERT2RND model in \citet{rothe2019leveraging}. We use the same hyperparameters as \citet{rothe2019leveraging}, which were selected after extensive tuning.

\textbf{\BertRnn} Instead of a transformer decoder, we use an RNN, specifically a gated recurrent neural network (GRU) \cite{cho2014learning}, as the decoder. Unlike the transformer decoder, which computes attention over the full input sequence, we do not use any attention mechanisms over the input to the decoder. Instead, we only pass the last hidden layer for the first token (``CLS'' token), forcing the model to encode all meaningful information about the input sequence into this single state. The RNN decoder, which consists of a single decoder layer, is substantially smaller than the transformer decoder used in the \BertTrans model.

\textbf{\BertRnnSc} Our third model uses the same architecture as the \BertRnn model, but we add an additional component to the loss function that encourages the start state and the end state of the decoder to be similar, which we call a state constraint (SC). The primary intuition behind this loss function is that it encourages the decoder to stay ``on topic'' while generating text, as it discourages the RNN from wandering too far away from where it started. It further encourages the start state to encode all information needed to generate the entire output sequence, rather than allowing the start state to focus on information in the beginning of the sequence and the end state to encode information for the end of the sequence.

The general form for the state of an RNN decoder \cite{cho2014learning} is
\begin{align}
    h_t = f(h_{t-1}, y_{t-1})
\end{align}
Here, $f$ is a GRU, $t$ $\in \{1, \dots, T\}$ is the target token position, and $h_{0}$ is initialized to the CLS token of the BERT source encoder.

The formula for the state constraint function is given in Equation \ref{eq:stateloss}: 
\begin{align}
d &= \frac{h_{0}}{||h_{0}||_2} - \frac{h_{T}}{||h_{T}||_2} \nonumber \\
{\mathcal{L}_{SC}} &= ||d||_2 \label{eq:stateloss}
\end{align}

The normalization terms force the loss term to focus on embedding direction rather than magnitude; they are necessary to account for the arbitrary magnitude of model states. During training, we multiply the state constraint loss, $\mathcal{L}_{SC}$, by a fixed scalar ($\alpha$) and add it to the standard cross-entropy ($\textit{CE}$) loss function. The final loss function is then given by:
\begin{align}
\mathcal{L} = \mathcal{L}_{CE} + \alpha\mathcal{L}_{SC} \nonumber
\end{align}

\textbf{\BertRnnAttn} Our final model also uses a transformer encoder and an RNN decoder. However, unlike the previous model, we pass the entire last layer of the encoder to the decoder and add an attention mechanism over this input sequence \cite{luong2015effective}. This model and the \BertTrans model are attention-based decoders, while the \BertRnn and the \BertRnnSc models do not use attention over the decoder input.

\section{Data}
\label{sec:data}

Our primary data set consists of articles from English Language Wikipedia collected on June 25, 2019. We filter out articles that contain the word ``redirect" and omit any section whose title has fewer than 2 characters. We extracted sections and section titles from each article and randomly divided the data into train, test, and development sets, using an 80/10/10 split (11.43M/1.43M/1.44M articles).

Wikipedia articles are often hierarchical, containing multiple subsections. However, we make no distinction between titles that are complete sections and titles that are subsections. This lack of distinction makes the generation task harder, as our models are not able to take advantage of hierarchical information and also allows our models and results to better generalize to other data sets that do not have this hierarchy.

More detailed statistics on the data set are shown in \Tref{tab:wiki_data}. For reference, we also show statistics for the commonly-used Gigaword Corpus \cite{rush2015neural}, which we also use to evaluate our models in \Sref{sec:results}. The Gigaword corpus entails an abstractive short summary generation task: given the first sentence of a newspaper article, predict the article title. We use this task for comparison because it uses a well-studied data set that is more similar to the Wikipedia section heading generation task than other text generation tasks, such as summarization tasks, which typically involve much longer outputs \cite{narayan2018don}. However, as shown in \Tref{tab:wiki_data}, there are notable differences between these data sets.

\begin{table}[hbt]
    \centering
    \begin{tabular}{lcc}
    \hline\hline
    & Wikipedia & Gigaword \\
    \hline
     Total size & 14.3M & 4.4M\\
     Train size & 11.43M & 4.2M \\
     Test size & 1.43M & 1.9K\\
     Dev size &  1.44M & 210K \\
    \hline
     Distinct titles & 45.25\% & 80.45\%  \\
     Unique titles & 41.82\%  & 70.28\%  \\
     Most common title & 3.35\% & 0.17\% \\
     Avg. words per title & 2.65 & 8.64 \\
     \end{tabular}
\caption{Overview of the Wikipedia section title data, as compared with the Gigaword corpus. ``Distinct titles'' refers to the total number of titles with duplicates removed. ``Unique titles'' refers to the number of titles that occur exactly 1 time. In general, the Wikipedia titles are shorter and more repetitive than Gigaword titles.}
\label{tab:wiki_data}
\end{table}

In the Wikipedia corpus, across 14.3M data points, there are only 6.5M distinct headings (45.25\% of all titles). Approximately 6M headings (41.82\%) occur only 1 time in the data, meaning the other 0.5M headings are reused multiple times across 8.3M articles to constitute the remainder of the corpus. The most common heading, ``History'', occurs 480K times in the data set, making up 3.35\% of the total corpus. Other common headings include ``Career'' (181K), ``Biography'' (151K), ``Early Life'' (111K), ``Background'' (102K) and ``Plot'' (96K).

In contrast, the titles in Gigaword are generally longer and more distinctive than the Wikipedia section titles, with 80.45\% of all titles being unique. However, in the absence of generic abstract headings like ``History'', the Gigaword corpus tends to be more extractive, meaning there is high token-overlap between articles and their titles. The Wikipedia corpus is also much larger than Gigaword, which facilitates analyses.

\section{Experimental Setup}
\label{sec:experimental_setup}

For all encoders, we use the BERT-base uncased model. Thus, we lowercase all text and use wordpiece tokenization from the public BERT wordpiece vocabulary \cite{devlin2019bert}. We use the same preprocessing pipeline, including word-piece tokenization, when computing target text length and extractive scores.

For all models, we limit the encoder input size to 128 tokens and the decoder output size to 32 tokens and use a batch size of 32. We generally use a learning rate of 0.05 with square root decay, 40K warm-up steps, and the Adam optimizer; however, for the RNN models with the Gigaword data, we use 100K warm-up steps, clip gradients to 20, and optimize with Adagrad, which we found to produce smoother training curves. For the state constraint models, we start by setting the scalar $\alpha = 0$, and linearly increase $\alpha$ to 1, between 100k and 200k training steps. We train the RNN models for 2M steps using v100 GPUs, and we train the transformer models for 500K steps using TPUs. In practice, we find that the RNN performance stops improving within 1M steps and the transformer performance stops within 50K steps.

\section{Results and Analysis}
\label{sec:results}

\subsection{Section Heading Generation}

Our main task is to generate a Wikipedia section title given the section text. \Tref{tab:res_wiki_sections} reports results using standard summarization metrics: Rouge-1, Rouge-L, and exact match. Rouge-1 measures the unigram overlap between the generated text and the reference text; Rouge-L measures the longest subsequence that occurs in both the generated text and the reference; exact match measures if the generated text exactly matches the reference. The \BertRnnAttn model performs the best overall. The \BertTrans and the \BertRnnSc models perform approximately the same, and both outperform the regular \BertRnn model.

Because the Wikipedia dataset contains diverse types of headings, including short abstractive headings and long extractive headings, we subdivide our test data in order to better understand model performance. In \Tref{tab:res_wiki_length}, we examine how well these models generate outputs of different lengths by dividing the test set according to the number of tokens in the target headings.

All of the RNN decoders outperform the transformer decoder for short headings containing 1-5 tokens, and the \BertRnnSc model performs the best overall. Over these short headings, the attention mechanism provides little advantage. However, the two attention-based decoders, \BertTrans and \BertRnnAttn outperform the RNNs without attention for mid-range-length headings containing 5-10 tokens, which is consistent with prior work suggesting that attention improves the modeling of long-term dependencies \cite{vaswani2017attention}. Nevertheless, on headings with $>10$ tokens, the Rouge-L scores for all decoders decline.

\begin{table}[ht]
    \centering
    \begin{tabular}{lccc}
    \hline\hline
   & Rouge-1 & Rouge-L & Exact \\
   \hline
   \BertTrans & 52.0 & 51.9 & 39.3 \\ %bert2trans_wiki3 1565749705
   \BertRnn & 50.2 & 50.1 & 33.8 \\ % bert2rnn_nosl_wiki1 1565587067
   \BertRnnSc & 52.6 & 52.4 & 36.5 \\ % bert2rnn_sl_wiki5 1565607262
   \BertRnnAttn & \textbf{54.4} & \textbf{54.3} & \textbf{40.5} \\ % bert2rnn_attn_wiki3 1566538264
    \end{tabular}
    \caption{Results on Wikipedia section heading generation over the full test set.}
    \label{tab:res_wiki_sections}
\end{table}

\begin{table}[ht]
    \centering
   \begin{tabular}{lcccc}
       \hline\hline
   \# Tokens & 1-5 & 5-10 & 10-15 & 15+ \\
   Data Size & 1M & 300K & 56K & 9K \\
    \hline
    \BertTrans & 52.5 & 53.8 & 36.3 & 20.8 \\
    \BertRnn & 54.0 & 39.6 & 35.1 & 25.3 \\
    \BertRnnSc & \textbf{55.8} & 44.2 & 37.7 & 24.0 \\
    \BertRnnAttn & 54.4 & \textbf{55.7} & \textbf{47.8} & \textbf{32.9} \\
    \end{tabular}
    \caption{Rouge-L on Wikipedia section heading generation by length. The attention-based decoders outperform the decoders without attention on target texts containing 5-10 tokens, but not on shorter target sequences.}
    \label{tab:res_wiki_length}
\end{table}

Prior work has also examined the trend of extractiveness in text generation models, specifically observing that models achieve high performance when they can copy input tokens directly into the output, rather than having to encode semantic information and produce new tokens \citep{nallapati2016abstractive,cheng2016neural,see2017get,nallapati2017summarunner,narayan2018don,grusky2018newsroom,pasunuru2018multi}. Because we ultimately extract embeddings from our models, understanding to what extent they copy tokens or encode more abstract information offers insight into how useful we can expect embeddings to be. To examine this, we introduce a metric called \textit{extractive score}, which measures what percentage of the output text can be directly copied from the input text:
$\frac{|T_{target} \bigcap T_{input}|}{|T_{target}|}$, where $T_{target}$ and $T_{input}$ represent the tokens in the target text and the input text respectively.

Thus, for a section and title pair, an extractive score of 0 indicates that there is no token overlap between the title and the section text, while a score of 1 indicates that every token in the title is also in the section text. Because of the short length of our section titles, we focus on unigrams, rather than examining higher-order n-grams. When computing extractive scores, we use the same text preprocessing pipeline as used in our models, including wordpiece tokenization and lowercasing.

\begin{figure}[ht]
    \centering
    \includegraphics[width=\linewidth]{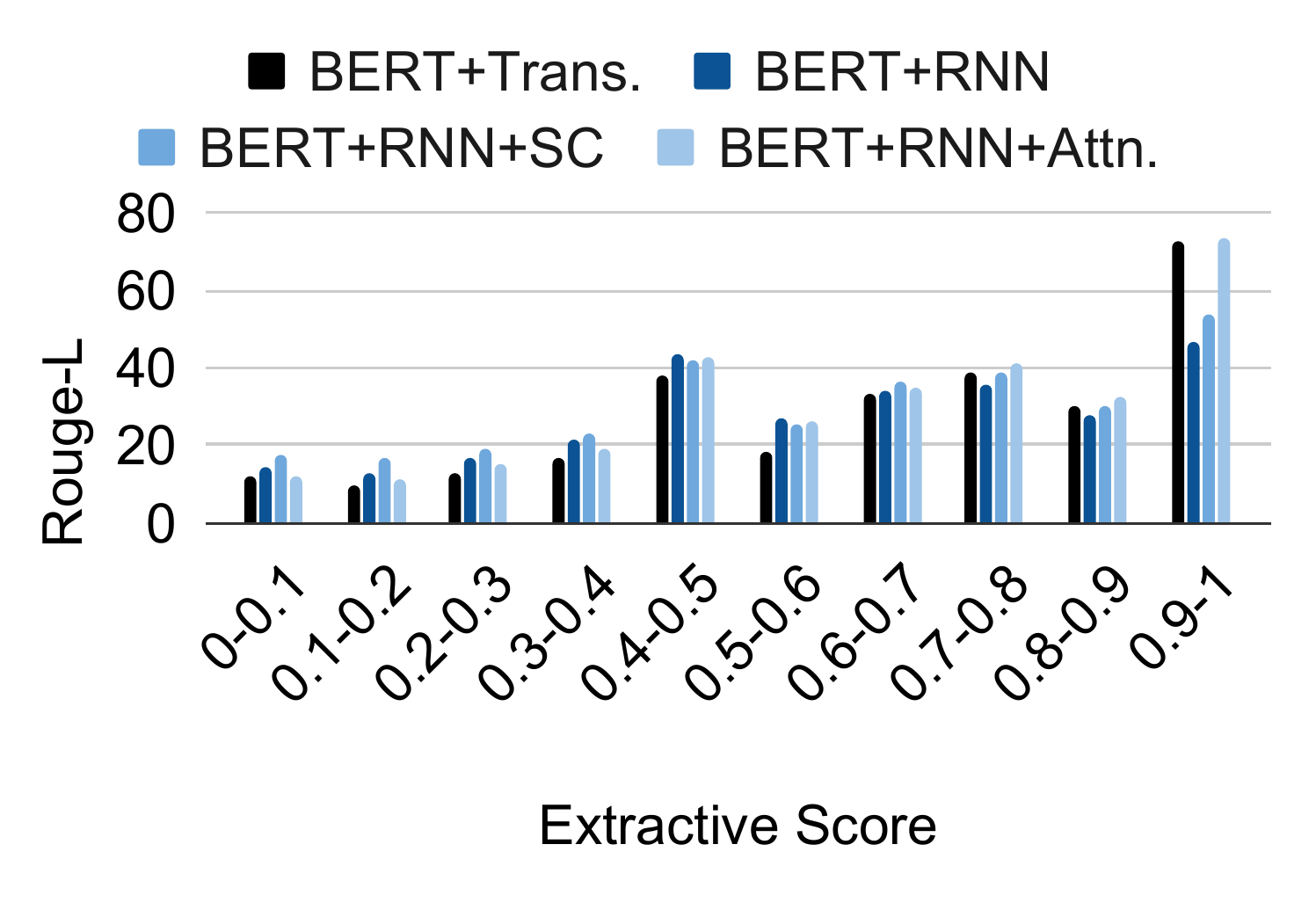}
    \caption{Rouge-L scores for each model over test data of length 5-10 tokens (300K test samples), segmented according to extractive score.}
    \label{fig:extractive_wiki}
\end{figure}

In \Fref{fig:extractive_wiki}, we limit the test data to headings with 5-10 tokens and divide it into segments according to extractive score. The \BertRnn and \BertRnnSc models outperform the attention-based models on data with low extractive scores ($\le 0.5$). The higher performance of the \BertTrans and the \BertRnnAttn models as compared to the \BertRnn and \BertRnnSc models over this data segment (\Tref{tab:res_wiki_length}) is almost entirely on headings where the extractive score is $\ge 0.9$. The attention-based models are not better at producing long titles \textit{in general}, but rather their ability to copy from the input text allows them to generate long titles \textit{when they are extractive}.

% \begin{table}[ht]
%    \centering
%    \begin{tabular}{lccc}
%    \hline\hline
%    & Partial Correlation \\
%    \hline
%   \BertTrans & 0.215 \\ %bert2trans_wiki3 1565749705
%   \BertRnn &  0.115 \\ % bert2rnn_nosl_wiki1 1565587067
%   \BertRnnSc & 0.136 \\ % bert2rnn_sl_wiki5 1565607262
%   \BertRnnAttn & 0.205 \\ % bert2rnn_attn_wiki3 1566538264
%    \end{tabular}
%    \caption{Partial correlations between Rouge-L and extractive score, controlled for length. All values are statistically significant.}
%    \label{tab:res_wiki_corr}
%\end{table}

\begin{table}[ht]
    \centering
    \begin{tabular}{cccc}
    \hline\hline
    \BertTrans & \BertRnn & \BertRnnSc & \BertRnnAttn \\

    \hline
     0.215 & 0.115 & 0.136 & 0.205 \\
    \end{tabular}
    \caption{Partial correlations between Rouge-L and extractive score, controlled for length. All values are statistically significant.}
    \label{tab:res_wiki_corr}
\end{table}

We can further examine this trend by computing correlations between Rouge-L and extractive score. However, as \Tref{tab:res_wiki_length} shows, all decoders perform differently over texts of different lengths. Thus, in order to isolate the effect of extractiveness, we compute partial correlations \cite{rummel1976understanding}. The idea behind a partial correlation is to identify the relationship between two variables $X$ and $Y$ that is not explained by a confound $Z$. We first compute the residuals $e_{X,i}$ and $e_{Y,i}$, and then compute the correlation between these residuals:

$${e_{X,i}=x_{i}-\langle \mathbf {w} _{X}^{*},\mathbf {z} _{i}\rangle }$$
$${e_{Y,i}=y_{i}-\langle \mathbf {w} _{Y}^{*},\mathbf {z} _{i}\rangle }$$
$$\texttt{Partial Correlation} = \rho_{e_{X,i},  e_{Y,i}}$$

where ${w} _{X}^{*}$ and ${w} _{Y}^{*}$ are the coefficients learned by a linear regression between $X$ and $Z$ and between $Y$ and $Z$. In our case, $X=$ Rouge-L, $Y=$ extractive score, and $Z=$ target length.

\Tref{tab:res_wiki_corr} reports results. For all models, the resulting correlations are positive, indicating that they generate extractive headings better than non-extractive headings. However, the correlations for the \BertTrans and \BertRnnAttn models are highest. Overall, these results suggest that decoders with attention mechanisms achieve high performance on this task because they better copy tokens from the input into the output, rather than because they encode more semantics.  Encoding semantic information is essential for generating section embeddings, which we extract and evaluate in Section~\ref{sec:clustering_results}.

\begin{table}[h]
    \centering
    \begin{tabular}{lccc}
    \hline\hline
    & Rouge-1 & R.-L & P. Corr  \\
    \hline
    \citet{song2019mass} & 38.7 & 36.0 & -- \\
    \hline
    \BertTrans & \textbf{37.1} & \textbf{34.6} & 0.647 \\
    \BertRnn & 35.6 & 32.6 & 0.619 \\
    \BertRnnSc & 35.1 & 32.8 & 0.630 \\
    \BertRnnAttn & 36.3 & 33.8 & 0.667 \\
    \end{tabular}
    \caption{Results on Gigaword heading generation. The correlations between extractive score and model performance are stronger than for the Wikipedia corpus for all models. All correlations are statistically significant.}
    \label{tab:res_giga_sections}
\end{table}

\begin{table*}[ht]
    \centering
    \begin{tabular}{lcccc}
    \hline\hline
     & Homogeneity & Completeness & V-measure & ARI  \\
     \hline
    Doc2Vec & 0.334 & 0.443 & 0.381 & 0.065  \\
    TF-IDF & 0.428 & 0.361 & 0.392 & 0.044 \\ % The small number is size 768 and the large is size 1000 but I don't think it matters
    \hline
    \BertTrans & 0.633 & 0.529 & 0.576 & 0.065 \\
    \BertRnn & 0.668 & 0.558 & 0.608 & 0.079 \\
    \BertRnnSc & \textbf{0.670} & \textbf{0.561} &  \textbf{0.611} & \textbf{0.088} \\
    \BertRnnAttn & 0.626 & 0.521 & 0.569 & 0.067 \\
    \end{tabular}
    \caption{Results on Wikipedia section clustering. The \BertRnnSc model performs the best on all metrics.}
    \label{tab:res_clustering}
\end{table*}

\subsection{Gigaword Results}
\label{sec:gigaword_results}

In order to compare our models against published benchmarks and to generalize our observations about extractiveness, we conduct the same experiments over the Gigaword corpus as the Wikipedia corpus, using the established train, test, and dev splits \cite{rush2015neural}.

\begin{figure}[ht]
    \centering
    \includegraphics[width=\linewidth]{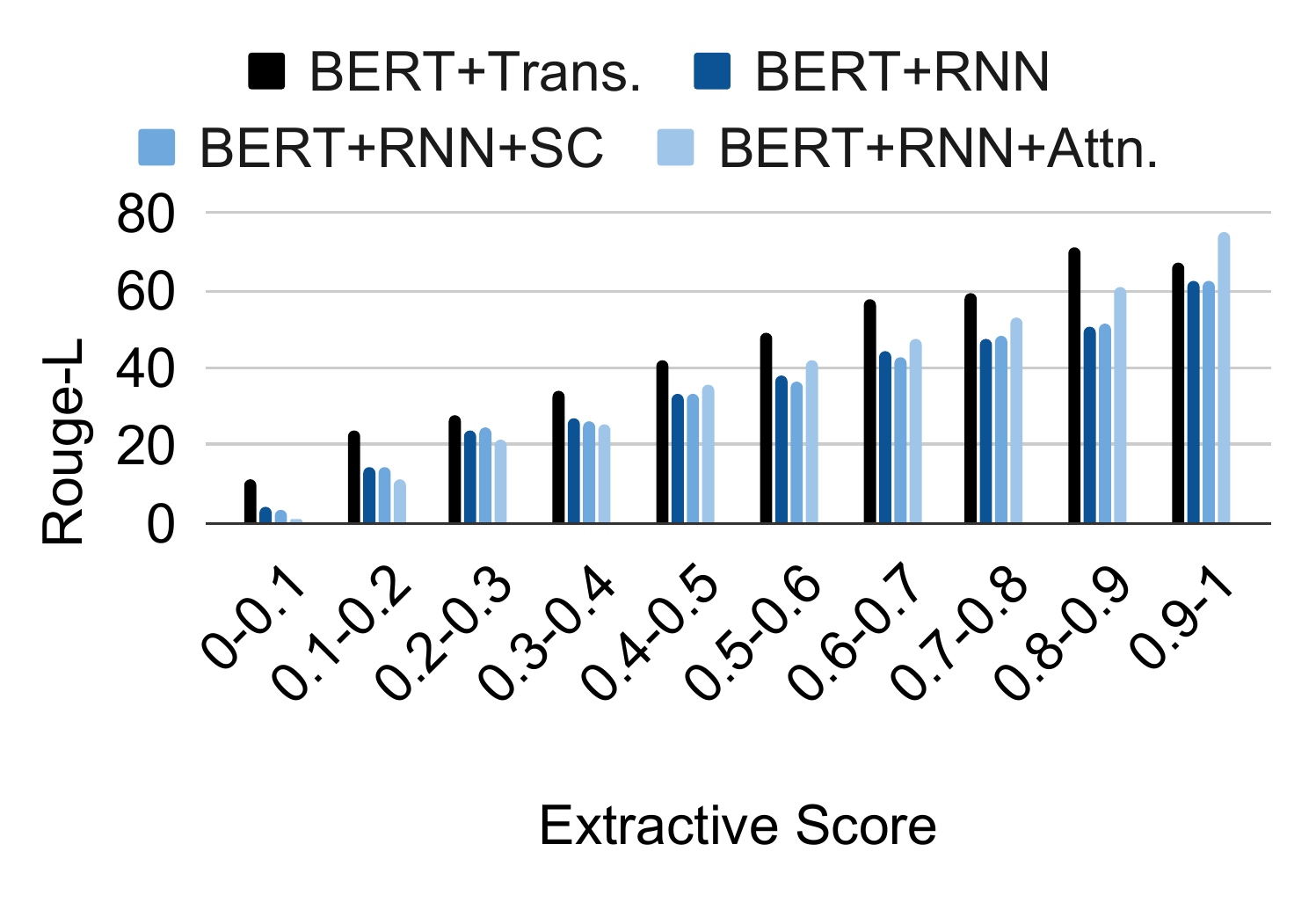}
    \caption{Rouge-L scores for each model over the Gigaword test data of length 5-10 tokens, segmented according to extractive score. Each data segment contains at least 35 samples.}
    \label{fig:extractive_giga}
\end{figure}

\Tref{tab:res_giga_sections} reports the results of our models as well as a state-of-the-art model for reference \cite{song2019mass}. Like \BertTrans, the MASS model from \citet{song2019mass} uses a transformer encoder-decoder architecture but with generalizations that allow for additional pre-training. From our models, the transformer decoder performs the best overall. However, the attention-based decoders \BertTrans and \BertRnnAttn also have the highest partial correlations, suggesting much of their performance stems from extractive titles. For all models the partial correlations between Rouge-L and extractive score are higher for the Gigaword corpus than for the Wikipedia corpus. This correlation is visually evident in \Fref{fig:extractive_giga}, which we constructed the same way as \Fref{fig:extractive_wiki}.

\Fref{fig:extractive_giga} mirrors the trend in the Wikipedia data (\Fref{fig:extractive_wiki}). While the \BertTrans model performs well across all extractiveness levels, the RNNs with and without attention perform similarly for lower levels of extractiveness. However, the \BertRnnAttn begins outperforming the RNNs without attention when the extractive score is $\ge0.5$, and especially when the extractive score is $\ge 0.9$.

\subsection{Section-embedding Generation and Clustering}
\label{sec:clustering_results}

While labeling sections can improve Wikipedia articles and identify the type of information contained in general paragraphs, embedding representations for paragraphs and documents can offer a more useful way to structure corpora, by facilitating information clustering and retrieval. Rather than creating generic all-purpose embeddings \cite{le2014distributed}, our generative models facilitate creating embeddings that target specific information, in our case, the title of the section.

We extract internal states from our models as section embeddings, and we evaluate them through a clustering task. Because many Wikipedia articles use the same generic headings, like ``History'' and ``Plot'', we can use these headings as gold cluster assignments by assuming that all sections with the same title constitute a cluster.

For all models, we use the final hidden layer for the first token in the input sequence (CLS token) as the embedding. In the case of the RNN decoder, this embedding is also the initial state of the RNN, and thus is the single state that the model is forced to encode the entire input sequence into.\footnote{For \BertTrans and \BertRnnAttn, preliminary experiments showed that using this hidden state as the  embedding achieved strictly better performance than other pooling possibilities.}

We cluster these embeddings using k-means clustering, where we set the number of clusters to the true number of clusters in the gold cluster assignments. We discard any section titles that occur fewer than 100 times, ensuring that the minimum size of any cluster is 100, resulting in 467,286 data points and 755 clusters. The large number of data points makes this task particularly difficult.

\Tref{tab:res_clustering} reports results using standard metrics for evaluating a proposed cluster assignment against gold data \cite{hubert1985comparing,rosenberg2007v}. Homogeneity assesses to what extent each cluster contains only members of the same class (e.g. does each cluster contain only sections with the same title?); completeness assesses to what extent members of the same class are in the same cluster (e.g. are sections with the same title in the same cluster?); V-measure is the harmonic mean between homogeneity and completeness; and adjusted Rand index (ARI) counts how many pairs of data points are assigned to the same or different clusters in the predicted and gold clusterings. On all metrics, the \BertRnnSc model performs the best.

To show how our embeddings, which are tailored to this task, differ from off-the-shelf embeddings, we report results using embeddings constructed from two popular methods for generating document embeddings: distributed representations using Doc2Vec \cite{le2014distributed,lau2016empirical,paraemb2019} and sparse embeddings using TF-IDF weighting \cite{banerjee2007clustering}. We train a Doc2Vec model over the training set using a window size of 5 and embedding size of 768, to match the embedding size of our models, and then infer embeddings over the test set. For the TF-IDF vectors, we give this method an additional advantage by directly training the model over the test set with an embedding size of 1000. As expected, all of our models outperform these off-the-shelf models.

Unlike off-the-shelf models, our customizable models encourage the embeddings to encode information specific to our prediction task. In this case, we train them to encode section title information. However, by training our models on a different prediction task, such as predicting the name of a newspaper outlet or a comment on a newspaper article, we can encourage the model to generate document embeddings that encode different information. Thus, our model architecture offers a way to generate high-quality document embeddings that encode information specific to the task at hand.

\section{Related Work}
While we introduce the task of Wikipedia section heading generation, the task of article headline generation using the Gigaword corpus has been well-studied, primarily using an encoder-decoder architecture with additional modules like attention or copy mechanisms \citep{rush2015neural,nallapati2016abstractive}. \citet{zhang2019pretraining} further explore how to leverage the pretrained BERT model for abstractive summarization, primarily using the CNN/Daily Mail data set. \citet{rothe2019leveraging} perform a comprehensive assessment of pretrained language models for text generation tasks, including the Gigaword task. Our \BertTrans model is identical to their BERT2RND model and achieves comparable results over the Gigaword corpus.

The high level of extraction in existing text generation tasks has motivated the use of mechanisms that explicitly copy input text into the output \cite{see2017get} or the introduction of new data sets \cite{narayan2018don,grusky2018newsroom}.  Furthermore, models trained for extractive summarization often outperform abstractive models on abstractive data sets \cite{cheng2016neural,nallapati2016abstractive,nallapati2017summarunner}. Our work extends these results by showing that even abstractive models are implicitly learning extraction, as they perform better on extractive text. Our metric for measuring extractiveness is similar to the `novel n-gram percentage' proposed by \citet{see2017get}; however, we use the same input pipeline for computing this metric as for training our models, and we correlate extractive score with performance, rather than just using it as an extrinsic measure of abstraction \cite{pasunuru2018multi}.

In our Wikipedia section heading generation task, the prevalence of generic headings makes the task more abstractive than datasets like Gigaword \cite{rush2015neural}, or even other short-text generation tasks, like email subject prediction \cite{zhang2019email}, which makes it a useful dataset for analyzing model performance. It is also extrinsically useful - most automated methods for improving Wikipedia focus on creating new content, such as through multi-document summarization \citep{liu2018generating} or generating text from structured data \citep{lebret2016neural}. However, less than 1\% of all English Wikipedia articles are considered to be of quality class “good”, suggesting there is a need for improving existing articles. \citet{piccardi2018structuring} show that many low quality articles consist of 0-1 sections and present a method for recommending new sections for an author to add to the article. Our approach offers a way to label existing paragraphs as distinct sections.

Our approach also results in document embeddings, which we show can be used to cluster sections. Document embeddings are useful for a variety of tasks including news clustering \citep{banerjee2007clustering,hu2009exploiting}, argument clustering \citep{reimers2019classification}, and as features for downstream tasks like text classification \cite{lau2016empirical,liu2018learning}. While TF-IDF vectors have historically been a popular construction method \cite{banerjee2007clustering}, more recent methods have focused on distributive representations, particularly Doc2Vec, a generalization of the Word2Vec algorithm \cite{le2014distributed,lau2016empirical,paraemb2019}.

Finally, the growing popularity of pretrained language models like BERT has led to numerous investigations on what these models learn \cite{liu2019linguistic,goldberg2019assessing,jawahar2019does}. Most investigations involve using targeted probing tasks. While our work shares similar goals, in that we investigate what type of information these models learn, we focus on data subsets and performance analysis.

\section{Future Work}

Our work offers several avenues for future exploration. We focus only on English Wikipedia. However, there are numerous language editions of Wikipedia, many of which have far fewer articles than the English edition and could benefit from tools for text generation.\footnote{\url{https://en.wikipedia.org/wiki/List_of_Wikipedias}} Additionally, while we discard the hierarchical nature of Wikipedia sections, this information could offer a way to improve model performance (potentially at the cost of generalizability to other data sets). Furthermore, while we evaluate the performance of our generated section embeddings for clustering, more work is needed to assess their usefulness on other tasks, such as retrieving relevant sections from a query, measuring section similarities, or as features for text classification.

\section{Conclusions}
Overall, our work introduces the task of generating section titles for text. We also introduce the \BertRnnSc model and demonstrate how RNN decoders can be utilized for short text generation and improved section embeddings. Specifically, our method for generating text embeddings, which involves leveraging internal states of models trained for generation, allows the embeddings to contain targeted information that maximizes their usefulness for specific tasks.

\section{Acknowledgements}
We would like to thank anonymous reviewers, Vidhisha Balakrishna, Keith Hall, Shan Jiang, Kevin Lin, Riley Matthews, and Yulia Tsvetkov for their helpful feedback and advice.

\bibliography{acl2020}
\bibliographystyle{acl_natbib}

% \break
% \setcounter{page}{1}
% \appendix

\end{document}